\def\year{2021}\relax
\documentclass[letterpaper]{article} 
\usepackage{aaai21}  
\usepackage{times}  
\usepackage{helvet} 
\usepackage{courier}  
\usepackage[hyphens]{url}  
\usepackage{graphicx} 
\usepackage{natbib}  
\usepackage{caption} 
\usepackage{bm}
\usepackage{amsmath}
\usepackage{amssymb}
\usepackage{bbm}

\usepackage{booktabs}

\usepackage{multirow}

\title{Time to Transfer: Predicting and Evaluating Machine-Human Chatting Handoff}
\author{
    Jiawei Liu,\textsuperscript{\rm 1}
    Zhe Gao,\textsuperscript{\rm 2}
    Yangyang Kang,\textsuperscript{\rm 2}
    Zhuoren Jiang,\textsuperscript{\rm 3}
    Guoxiu He,\textsuperscript{\rm 1}
    \\
    Changlong Sun,\textsuperscript{\rm 2,3}
    Xiaozhong Liu,\textsuperscript{\rm 4}\thanks{Corresponding authors.}
    Wei Lu\textsuperscript{\rm 1}\footnotemark[1]
    \\
}
\affiliations{
    \textsuperscript{\rm 1}Wuhan University, Wuhan, China\\
    \textsuperscript{\rm 2}Alibaba Group, China\\
    \textsuperscript{\rm 3}Zhejiang University, Hangzhou, China\\
    \textsuperscript{\rm 4}Indiana University Bloomington, Bloomington, United States\\

    \{laujames2017, guoxiu.he, weilu\}@whu.edu.cn, \{gaozhe.gz, yangyang.kangyy\}@alibaba-inc.com, \\
    jiangzhuoren@zju.edu.cn, changlong.scl@taobao.com, liu237@indiana.edu

}
\begin{document}
\maketitle
\begin{abstract}
Is chatbot able to completely replace the human agent? The short answer could be – ``\textit{it depends}...''. For some challenging cases, e.g., dialogue's topical spectrum spreads beyond the training corpus coverage, the chatbot may malfunction and return unsatisfied utterances. This problem can be addressed by introducing the Machine-Human Chatting Handoff (MHCH) which enables human-algorithm collaboration. To detect the \textit{normal/transferable} utterances, we propose a Difficulty-Assisted Matching Inference (DAMI) network, utilizing difficulty-assisted encoding to enhance the representations of utterances. Moreover, a matching inference mechanism is introduced to capture the contextual matching features. A new evaluation metric, Golden Transfer within Tolerance (GT-T), is proposed to assess the performance by considering the tolerance property of the MHCH. To provide insights into the task and validate the proposed model, we collect two new datasets. Extensive experimental results are presented and contrasted against a series of baseline models to demonstrate the efficacy of our model on MHCH.
\end{abstract}

\section{Introduction}
\label{sec:intro}
\begin{figure}[t]
\centering
  \includegraphics[width=3.2in]{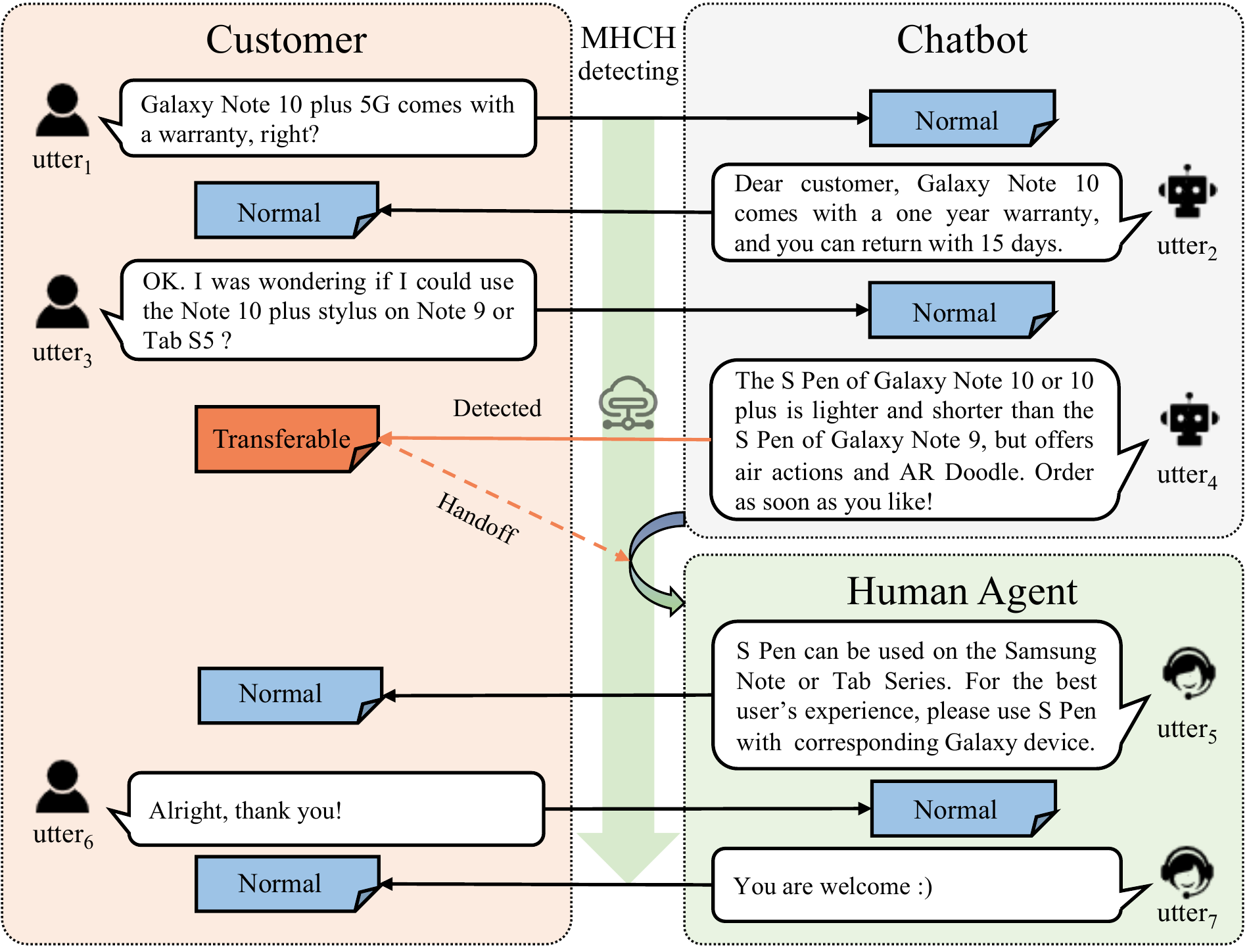} 
  \caption{A dialogue example in sales customer service, where orange transferable label denotes this dialogue session needs handoff from chatbot to human agent.}
  \label{dialog_example} 
\end{figure}

Chatbot, as one of the recent successful applications in artificial intelligence, makes many people believe that it will be able to replace the human agent in the foreseeable future. Deep learning efforts from industry and academia boost the development of chatbot and have rewarded many inspiring accomplishments \cite{high2012era,zhou2020design,liu2020you}. However, Morar Consulting's study\footnote{https://www.mycustomer.com/service/channels/could-chatbots-ever-completely-replace-human-agents} suggests that only 21\% respondents said they would like to be able to purchase goods and services from chatbots. Besides, due to the complexity of human conversation, automatic chatbots are not yet able to meet all the needs of users. The unsatisfied responses from algorithms may deteriorate user experience and cause business failure \cite{radziwill2017evaluating}. While scholars can consistently explore new models to further enhance the performance, in this study, we try to solve this problem from a perspective by combing chatbots with human agents. Specifically, we target the \textbf{Machine-Human Chatting Handoff (MHCH)}, which aims to enable smooth and timely handoff from chatbot to human agent.

Figure \ref{dialog_example} depicts an exemplar MHCH case of customer service for online sales in an E-commerce ecosystem. While chatbot successfully answers the first question from user (utter$_1$ and utter$_2$), unfortunately, it fails to address the second one decently (utter$_3$ and utter$_4$). In the MHCH task, the model should consistently monitor the progress of conversations, be able to predict the chatbot failures (potential algorithm failures, negative emotions from the user, etc.), and transfer the conversations to the human agent (a.k.a handoff) to ensure the smooth dialogue. Recently, there are already several works focus on the human-machine cooperation for chatbots. For instance, Huang, Chang, and Bigham \shortcite{huang2018evorus} integrated crowds with multiple chatbots and voting systems. Rajendran, Ganhotra, and Polymenakos \shortcite{rajendran2019learning} paid attention to transfer conversations to human agents once encountered new user behaviors. Different from these prior tasks, the MHCH task aims to sequentially label each utterance in the dialogue as \textit{normal} or \textit{transferable} based on the dialogue content and its contextual information. The MHCH task can improve the ability of chatbot from two aspects. First, it can predict the potential risks of the dialogue and improve user satisfaction. Second, it helps to optimize the human agent resource allocation and reduce the business cost.

In this study, we propose a novel Difficulty-Assisted Matching Inference (DAMI) model for the MHCH task. The DAMI model is a hierarchical network that mainly consists of a difficulty-assisted encoding module and a matching inference mechanism. Considering the difficulty of the utterances affects the chance of handoff, we propose the difficulty-assisted encoding module. Besides the lexical and semantic information, it encapsulates lengths, part-of-speech tags, and term frequencies to learn and generate comprehensive representations of utterances. The matching inference module utilizes the contextual interaction to capture irrelevant or repeated utterances that may deteriorate user experience. The classification is based on the contextual representations and the integration of the utterance representations and matching features. 

As for validating model performance for the MHCH task, none of the existing evaluation metrics seem to be very suitable because of the following reasons. \textbf{Imbalance}: The portion of \textit{normal} utterances and \textit{transferable} utterances is quite imbalanced. Conventional metrics, such as \textit{accuracy}, will be problematic \cite{soda2011multi}. \textbf{Comprehensiveness}: Correctness of detecting the \textit{transferable} utterances and coverage of \textit{transferable} utterances are both important. Neither \textit{precision} nor \textit{recall} can meet this requirement. Moreover, the contribution of prediction correctness for the dialogues without handoff cannot be ignored. \textbf{Tolerance}: Although \textit{F score}, \textit{AUC}, and other similar metrics are designed for the comprehensive evaluation, they can be too rigid for this task. For instance, a slightly earlier handoff is acceptable and could not be considered as a complete mistake. Inspired by \textit{the Three-Sigma Rule} \cite{pukelsheim1994three}, we propose a novel evaluation metric \textbf{Golden Transfer within Tolerance (GT-T)} for the MHCH task, which takes the imbalance, comprehensiveness, and tolerance into consideration.

To summarise, our main contributions are three folds: (1) We propose a novel DAMI model utilizing difficulty-related and matching information for the MHCH task; (2) We propose a new evaluation metric GT-T to cope with the imbalance, comprehensiveness, and tolerance problem in the MHCH task; (3) To assist other scholars reproduce the experiment outcomes and further investigate this novel but important problem, two real-world customer service dialogue datasets are collected, labeled\footnote{https://github.com/WeijiaLau/MHCH-DAMI}. The experimental results demonstrate the superiority of our approach.

\section{Related Works}
\label{sec:related}

In MHCH, a \textit{transferable} or \textit{normal} label is sequentially assigned to each utterance in a dialogue. To this end, the MHCH task can be defined as a supervised problem such as a classification problem or a sequence labeling problem.

Classic classification algorithms, such as LSTM \cite{hochreiter1997long} and TextCNN \cite{kim2014convolutional}, learn the representation of each utterance and predict the label straightforwardly. Recently, pre-trained language representations, such as BERT \cite{kenton2019bert}, have improved many NLP downstream tasks by providing rich semantic features. However, for instance, in a customer service scenario, customer's satisfaction can change gently with the dialogue moving forward. A decent model for the MHCH task should be able to sense user's emotional change or dissatisfaction in the earliest step, and a delayed handoff may cause user's impatience and abandonment. Compared to classic text classification methods, Yang et al. \shortcite{yang2016hierarchical}, Chen et al. \shortcite{chen2018dialogue}, Raheja et al. \shortcite{raheja2019dialogue}, and Dai et al. \shortcite{dai2020local} proposed different attention mechanisms to capture contextual information or dependencies among labels. Besides attention mechanisms, Yu et al. \shortcite{yu2019modeling} presented an adapted convolutional recurrent neural network that models the interactions between utterances of long-range context. Majumder et al. \shortcite{majumder2019dialoguernn} proposed a method based on recurrent neural networks that keeps track of the individual party states throughout the conversation and uses this information for emotion classification.

The mismatched or even irrelevant answers in some cases deteriorate the customer experience \cite{radziwill2017evaluating}. How to capture the matched or mismatched information have been widely studied in response selection of IR-based chatbots or sentiment analysis \cite{chen2017enhanced,liu2018multi,shen2018sentiment,mao2019multi}.
Besides users' emotions, text difficulty or readability affects the chance of handoff. There are a large number of researches based on machine learning methods to evaluate text difficulty \cite{collins2004language,hancke2012readability,jiang2018enriching}. Generally speaking, it is 
mainly related to the word (frequency, length, etc.), grammar, syntax, sentence length, etc. \cite{nelson2012measures,benjamin2012reconstructing}. Liao, Srivastava, and Kapanipathi \shortcite{liao2017measure} proposed a data-driven method to calculate the dialogue complexity and optimize the dispatching of human agents.

Since fully-automated chatbots still have limitations in the foreseeable future, a promising direction is to combine chatbots with human agents.
Earlier research, the crowd-powered conversational assist architecture, Evorus \cite{huang2018evorus}, integrated crowds with multiple chatbots and voting systems. Rajendran, Ganhotra, and Polymenakos \shortcite{rajendran2019learning} paid attention to transfer dialogues to human agents once encountered new user behaviors. After a period of deployment, they could automate itself with decent conversation quality. Different from them, we mainly focus on detecting transferable utterances which are one of the keys to improve the user satisfaction and optimize the human agent resource allocation.

\begin{figure*}[!ht]
\centering
\includegraphics[width=5.5in]{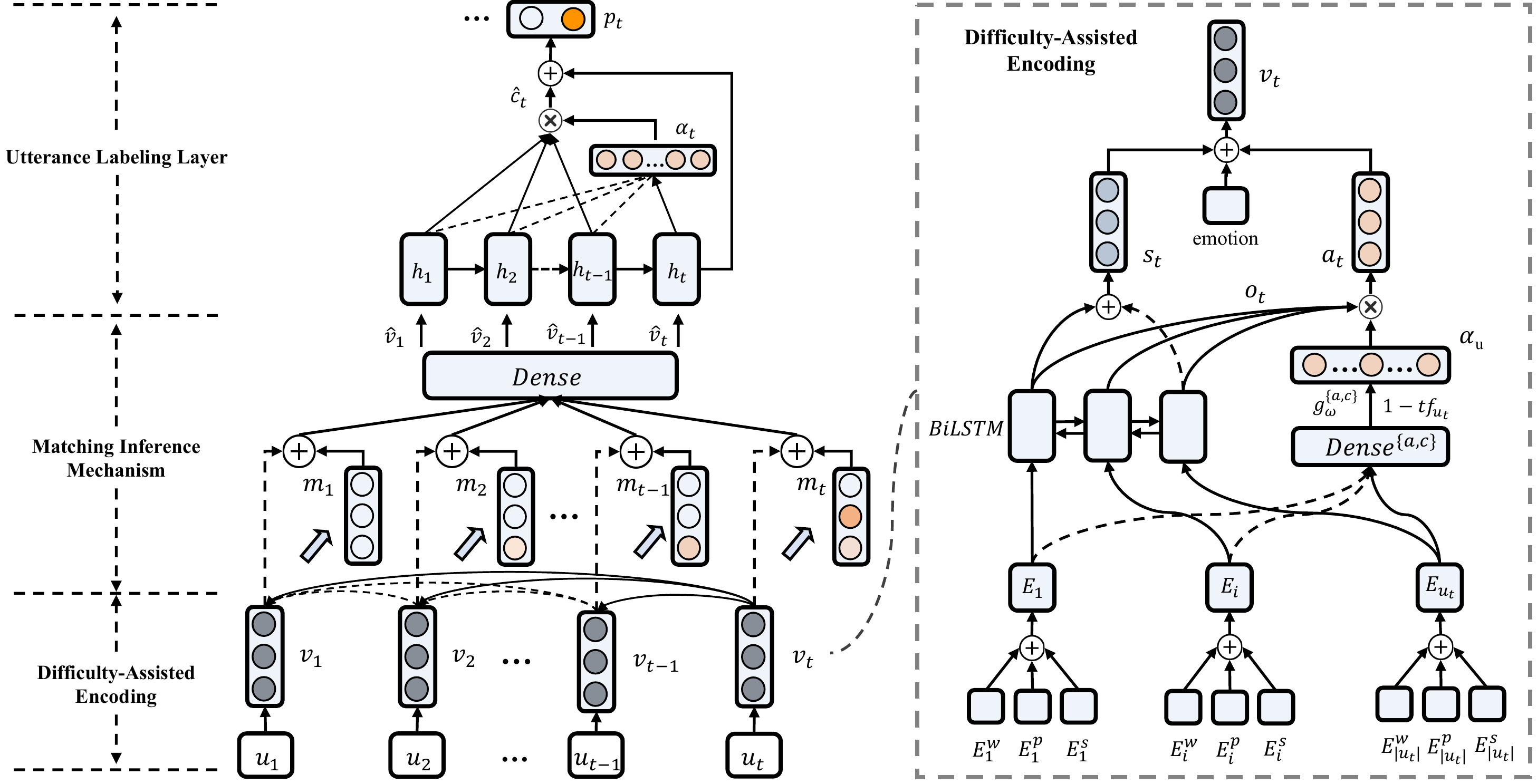} 
\caption{Overall architecture of Difficulty-Assisted Matching Inference (DAMI) network.}
\label{fig:dami}
\end{figure*}
\section{Methodology}

\subsection{Our Model}
Formally, we assume that a dialogue $D$ is composed of a sequence of utterances ${[u_1, ..., u_t, ..., u_L ]}$ with corresponding handoff labels ${[y_1, ..., y_t, ..., y_L ]}$, where ${y_t \in \mathcal{Y}}$. $\mathcal{Y}$ is the set of handoff labels (\textit{normal} and \textit{transferable}). \textit{Transferable} indicates the dialogue should be transferred to the human agent, whereas \textit{normal} indicates there is no need to transfer.
As shown in Figure \ref{fig:dami}, our model mainly consists of three components: \textit{Difficulty-Assisted Encoding}, \textit{Matching Inference Mechanism}, and \textit{Utterance Labeling Layer}.

\textbf{Difficulty-Assisted Encoding}: The difficulty of the utterance affects the chance of handoff. On the one hand, the chatbot may have difficulty to understand and answer the questions raised by customers. On the other hand, customers may have difficulty to understand or capture the useful information of the compound or lengthy responses given by the chatbot. Remarkably, the difficulty or readability of the text is related to the word (frequency, length, etc.), grammar, syntax, sentence length, etc. \cite{benjamin2012reconstructing}. Thus, besides the lexical and semantic information, we make use of positions, term frequencies, and part-of-speech tags to enhance the representation of utterance difficulty.

Suppose ${{u}_{t}=[ w_1, ..., w_{\left | u_t\right |}]}$ represents a sequence of words in ${u_t}$. These words are mapped into corresponding word embeddings ${\boldsymbol{E}_{u_t}^{w}\in \mathbb{R}^{d\times \left | u_t\right |}}$, where $d$ is the word embedding dimension.
Positional embedding \cite{vaswani2017attention}, which can encode order dependencies, is utilized to represent the increasing length information. ${\boldsymbol{E}_{1}^{p}\in \mathbb{R}^{d}}$ is the positional embedding of $w_{1}$ and ${\boldsymbol{E}_{u_t}^{p}\in \mathbb{R}^{d\times \left | u_t\right |}}$ is the positional embeddings list of words in $u_t$.
The one-hot representations of part-of-speech tags ${{\boldsymbol{E}_{u_t}^{s}\in \mathbb{R}^{n\times \left | u_t\right |}}}$ of words in ${u_t}$ are also generated. Note that $n$ is the number of part-of-speech tag categories. Hence, the overall embedding ${\mathbf{E}_{u_t} \in \mathbb{R}^{(2d+n)\times|u_t|}}$ of $u_t$ is:
\begin{align}
{\boldsymbol{E}_{u_t} = \boldsymbol{E}_{u_t}^{w} \oplus \boldsymbol{E}_{u_t}^{p} \oplus \boldsymbol{E}_{u_t}^{s}}
\end{align}
where $\oplus$ denotes concatenation operation. We utilize a bidirectional LSTM (BiLSTM) for obtaining the output of every time step $\boldsymbol{o}_{t} \in \mathbb{R}^{{2k} \times {\left | u_t\right |}}$ and  the last state concatenation of forward and backward $s_{t}\in \mathbb{R}^{2k}$, where $k$ is the number of hidden units of LSTM cell.
Briefly, we have:
\begin{align}
{\boldsymbol{o}_{t}, {s}_{t} = \text{BiLSTM}(\boldsymbol{E}_{u_t})}
\end{align}

In sales customer service dialogue, the roles of different participants would exhibit different characteristics \cite{song2019using}. Customers tend to use more succinct phrases, whereas agents tend to use more elaborated sentences \cite{liao2017measure}. Thus, we use different vectors to capture different roles' difficulty features:
\begin{align}
\textbf{\textit{W}}^{r}_{\omega} &= r_t\textbf{\textit{W}}^{c}_{\omega}+({1-r_t})\textbf{\textit{W}}^{a}_{\omega} \\
b_{\omega}^{r} &= r_t b_{\omega}^{c}+({1-r_t})b_{\omega}^{a} \\
\textbf{\textit{g}}^{r}_{\omega} &= r_t\textbf{\textit{g}}^{c}_{\omega}+({1-r_t})\textbf{\textit{g}}^{a}_{\omega}
\end{align}
where $\textbf{\textit{W}}_{\omega}^{\{ a,c \}} \in \mathbb{R} ^{ z \times d}$, $b_{\omega}^{\{a,c\}} \in \mathbb{R} ^{z}$ and $\boldsymbol{g}_{\omega}^{\{a,c\}} \in \mathbb{R} ^{z \times \left | u_t\right | } $ are trainable parameters shared across utterances of agents and customers, respectively. $z$ is the attention size. Since there are two roles in the MHCH task, we use $r_t \in \{0 ,1\}$ represents the role of utterance $u_t$.
Intuitively, the word frequency can also reflect the text difficulty: rarely used words (low word frequency) usually increase the understanding difficulty \cite{chen2018word}. Thus, we employ a term-frequency-adjusted attention mechanism to further enhance the difficulty-assisted representation $a_{t}\in \mathbb{R}^{2k}$:
\begin{align}
\boldsymbol{g}_{u_t} &= \text{tanh}(\textbf{\textit{W}}^{r}_{\omega} \boldsymbol{E}_{u_t} + b_{\omega}^{r}) \\
\boldsymbol{\alpha}_{u_t} &= {\text{softmax}((1- \text{tf}_{u_t})\cdot\boldsymbol{g}_{u_t}^{\top}\cdot \boldsymbol{g}_{\omega}^{r})} \\
a_t &= \sum\nolimits_{t=1}^{|u_t|}  \mathbf{o}_t \cdot \boldsymbol{\alpha}_{u_t}
\label{formular: alpha_u}
\end{align}
where $\boldsymbol{g}_{u_t} \in \mathbb{R} ^{z \times \left | u_t\right | }$ is the hidden representations and encoded by a fully connected layer. Then we compute the normalized difficulty weights $\boldsymbol{\alpha}_{u_t}$ with a local vector $\boldsymbol{g}_{\omega}^{r} $ and normalized term frequencies $\text{{tf}}_{u_t} \in [0, 1]$ of utterance $u_t$. 

Meanwhile, the customer's emotion is an indicator of service quality, which is potentially helpful for the handoff judgment. Hence, we also calculate an emotional polarity score $e_{t} \in \mathbb{R}$ for utterance $u_t$ by SnowNLP\footnote{https://github.com/isnowfy/snownlp}.

Finally, we get the utterance representation vector $v_{t} \in \mathbb{R} ^{K}$ (${K=4k+1}$) of $u_t$ as follow:
\begin{align}
    v_{t} = {s}_{t} \oplus {a}_{t} \oplus e_{t}
\label{eq:vt}
\end{align}

\textbf{Matching Inference Mechanism}:
The customer tends to turn to a human agent for help when the chatbot gives an unsatisfactory or irrelevant response. Practically, it is very difficult and paradoxical to directly judge whether the given answer is satisfactory. Nevertheless, the repeated utterances (the same responses or semantically similar questions) would increase customer's dissatisfaction. Based on this observation, we make the present utterance interact with preceding utterances to get the contextual matching features. For utterance $u_t$, we calculate its matching features $m_{t}$ by:
\begin{gather}
    {{m}_{t} = {v}^{\top}_t [{v}_{1},{v}_{2},...,{v}_{t-1}]}
\end{gather}

By masking out the future information of current utterance, the matching features of dialogue $D$ can be represented as ${\boldsymbol{V}_{M} = [m_1,..., m_t,...,m_L]}$, where ${\boldsymbol{V}_{M} \in \mathbb{R}^{L\times L}}$ is a lower triangular matrix with the diagonal value removed.
Then we use a fully connected layer to further integrate the matching features with utterance vector $\hat{v}_{t} \in \mathbb{R}^{k}$:
\begin{gather}
    {\hat{v}_{t} =  \text{ReLU}( \textbf{\textit{W}}_{\tau }(m_t \oplus v_t) + b_{\tau }) }
\end{gather}
where $\textbf{\textit{W}}_{\tau } \in \mathbb{R}^{k \times (K + T)}$ and $b_{\tau } \in \mathbb{R}^{k}$. $\text{ReLU}$ denotes the rectified linear unit activation \cite{nair2010rectified}.

\textbf{Utterances Labeling}:
The tendency of handoff also depends on the dialogue context. Thus, to connect the information flow in the dialogue, we feed the integrated representations into an LSTM:
\begin{gather}
    {h_t = \text{LSTM}(\hat{v}_{t}, h_{t-1}, c_{t-1})}
\end{gather}
where $h_t \in \mathbb{R}^{k}$ is the hidden state of LSTM for the utterance $u_t$ and ${c_{t-1}}$ is the memory cell state at time-step ${t-1}$.

Furthermore, to estimate a long-range dependency, we employ an attention mechanism \cite{majumder2019dialoguernn} to compute the context representation $\hat{c}_t \in \mathbb{R}^{k}$ of utterance $u_t$:
\begin{align}
    {\alpha}_{t} &= {\text{softmax}({h}^{\top}_t \textbf{\textit{W}}_{\alpha}  [h_1, h_2, ..., h_{t-1}])}\\
    \hat{c}_t &= {{\alpha}_{t} [h_1, h_2, ..., h_{t-1}]^{\top}}
\end{align}
where $[h_1, h_2, ..., h_{t-1}]$ are the hidden states for the preceding utterances of utterance $u_t$ and $\textbf{\textit{W}}_{\alpha} \in \mathbb{R}^{k \times k}$. Then we utilize a fully connected layer to get the context-aware representation $\hat{h}_{t} \in \mathbb{R}^{k}$:
\begin{gather}
    \hat{h}_{t} =  \text{ReLU}( \textbf{\textit{W}}_{\varsigma }(h_t \oplus \hat{c}_t) + b_{\varsigma }) 
\end{gather}
where  $\textbf{\textit{W}}_{\varsigma } \in \mathbb{R}^{k \times 2k}$ and $b_{\varsigma } \in \mathbb{R}^{k}$.

Since there are no dependencies among labels, we simply use a softmax classifier for performing handoff prediction:
\begin{gather}
    p_t = \text{softmax}(\textbf{\textit{W}}_{\gamma} \hat{h}_t + b_{\gamma}) 
\end{gather}
where $\textbf{\textit{W}}_{\gamma} \in \mathbb{R}^{2 \times k}$ and $b_{\gamma} \in \mathbb{R}^{2}$. $p_t \in \mathbb{R}^2$ is the predicted handoff probability distribution of $u_t$.

\textbf{End-to-End Training}:
We use categorical cross-entropy as the measure of loss ($\mathcal{L}$) during training:
\begin{align}
\mathcal{L} = - \frac{1}{I} \sum_{i=1}^{I}\sum_{t=1}^{L}\log p_{i,t}(y_{i,t}|D_{i}, \Theta ) +\frac{\delta }{2}\left \| \Theta \right \|_2^2
\end{align}
where $I$ is the number of dialogues, $D_i$ is the $i$-th dialogue, $p_{i,t}$ is the probability distribution of handoff label of $t$-th utterance of $D_i$, $y_{i,t}$ is the expected class label of $t$-th utterance of $D_i$, $\delta$ denotes the $L_2$ regularization weight and $\Theta$ denotes all the trainable parameters of model. 
We use backpropagation to compute the gradients of the parameters, and update them with Adam \cite{kingma2014adam} optimizer. 

\subsection{Golden Transfer within Tolerance}
For the MHCH tasks, it is vital to detect all of the \textit{transferable} utterances. Meanwhile, predicting \textit{normal} utterances as \textit{transferable} will also cause a waste of human resources. In order to comprehensively measure the model performance for predicting the \textit{transferable} or \textit{normal} label of an utterance, we adopt F1, Macro F1, and AUC.
However, the evaluation metrics like F1, Macro F1, and AUC have several limitations. First, if the predicted \textit{transferable} utterance is not the true \textit{transferable} utterance, these metrics will treat it as an error, no matter how close these two utterances are. This characteristic can be too rigid for the MHCH task. Second, F1 and AUC have no reward for the correct prediction of the dialogues without handoff.

To address above problems, we propose a new evaluation metric, namely \textbf{Golden Transfer within Tolerance (GT-T)}. Inspired by \textit{the Three-Sigma Rule} \cite{pukelsheim1994three}, the proposed GT-T allows a ``biased'' prediction within the tolerance range ${T \in \mathbb{N}}$. Specifically, 
in this paper, we set ${T}$ to range from 1 to 3 corresponding to GT-I, GT-II, and GT-III.

Suppose a dialogue session contains ${L \in \mathbb{N}}$ utterances and ${N \in \mathbb{N}}$ true \textit{transferable} utterances, where ${N \leq L}$. The number of predicted \textit{transferable} utterances is ${M \in \mathbb{N}}$, where ${M \leq L}$. The set of true \textit{transferable} utterances' positions is ${Q=\{ q_1, q_2, ...,q_{N}\}}$, where ${ 0 \leq q_j \leq {L-1} }$ and ${ j\in\{1,2,...,N\}}$. The set of predicted \textit{transferable} utterances' positions is ${P=\{p_1, p_2, ...,p_{M}\}}$, where ${0 \leq p_i  \leq {L-1}}$ and ${ i \in\{1,2,...,M\}}$. We compute this session's GT-T score as:
\begin{align}
\begin{cases}
 &0 ,{\text{ if }  {N = 0} , {M > 0} \text{ or } N>0, M=0} \\ 
 &1 ,{\text{ if } N=0, M=0} \\ 
& {\frac{1}{M} \sum\limits_{i=1}^{M} \mathop{\max}\limits_{1\le j\le N} exp \left(\frac{1}{\lambda \cdot sgn(\Delta_{ij}) - 1} \cdot \frac{\Delta_{ij}^2}{2(T+\epsilon)^2} \right) }, \text{others}
\end{cases}
\end{align}
Note that $\epsilon$ is an infinitesimal, $sgn$ represents the signum function and ${\Delta_{ij}=p_i - q_j}$. ${\lambda \in (-1, 1)}$ is an adjustment coefficient for the early handoff and the delayed handoff. An early handoff may improve the user experience but exacerbate the human agent resource scarcity. In turn, a delayed handoff may mitigate the human agent resource scarcity but sacrifice the user experience. ${\lambda > 0}$ means that the user experience is more important than the human agent resource. ${\lambda < 0}$ means that the human agent resource is more important than the user experience.

There are three advantages of the proposed GT-T metric:
(1) \textbf{Focusing on handoff.} Since the \textit{transferable} utterances are a minority in the dialogue with handoff (refer to Table \ref{tab:statistic}), conventional metrics, such as accuracy, will be problematic as the \textbf{Imbalance} of classes. Contrarily, GT-T mainly focuses on \textit{transferable}. Thus it will not be affected by \textit{normal} utterances.
(2) \textbf{Comprehensiveness.} For the MHCH task, the correctness of detecting the \textit{transferable} utterances and coverage of all \textit{transferable} utterances are both important. Moreover, handoff does not necessarily happen in all cases, which means the correct judgment of the dialogues without handoff should be rewarded.
(3) \textbf{Tolerance.} Although \textit{F score, AUC,} and other similar metrics are designed to comprehensively measure the model, they can be too rigid for the MHCH task. For example, if there is a dialogue with true labels sequences ${Q_0 = [0, 0, 0, 0, 0, 1]}$, it means that the dialogue needs to be transferred to the human agent at the sixth round. The F1, Macro-F1, AUC, GT-I, GT-II, and GT-III all scored 1 for prediction ${P_1 = [0, 0, 0, 0, 0, 1]}$. If there are enough human agents resources or the customer's patience is poor, the prediction, ${P_2 = [0, 0, 0, 0, 1, 0]}$, can be acceptable and shouldn't be considered as a complete mistake. In such situation, GT-I, GT-II, and GT-III scores (${\lambda=0}$) of ${P_2}$ are 0.61, 0.88, and 0.95, while the F1, Macro-F1, and AUC scores of ${P_2}$ are 0, 0.4, and 0.4. 
Obviously, GT-T is more suitable for MHCH task.

\section{Datasets}
\label{sec:data}
To the best of our knowledge, there is no publicly available dataset for the MHCH task. To address this problem, we propose two Chinese sales customer service dialogue datasets, namely \textbf{Clothing} and \textbf{Makeup}, which are collected from Taobao\footnote{https://www.taobao.com/}, one of the largest decentralized E-commerce platforms in the world.
Clothing is a corpus with 3,500 dialogues in the clothing domain and Makeup is a corpus with 4,000 dialogues in the makeup domain. User information is removed from both datasets. Each dialogue was annotated by an expert, and at least half of the dialogues were checked randomly by another expert. Five experts participated in the annotation work. Every utterance of the dialogue was assigned a \textit{transferable} or \textit{normal} label precisely without considering early or delayed handoff. Since the human agent is not involved in the dialogue, there may be multiple transferable utterances in each dialogue. Besides the \textbf{explicit demand} for human agent from the customer, an utterance will also be annotated as \textit{transferable} if the customer receives an \textbf{unsatisfactory answer} or expresses \textbf{negative emotions}. When the customer or chatbot repeats semantically similar utterances, the \textbf{repeated utterance} may also be annotated as \textit{transferable}.
\begin{table}[ht]
  \centering
  \footnotesize
    \begin{tabular}{lcc}
    \toprule
    \textbf{Statistics items} & \textbf{Clothing} & \textbf{Makeup} \\
    \midrule
    \midrule
    \# Dialogues & 3,500 & 4,000 \\
    \# Normal dialogues & 274   & 255 \\
    \# Dialogues with handoff  & 3,226 & 3,745 \\
    \# Utterances & 35,614 & 39,934 \\
    \# Normal utterances & 28,901 & 32,488 \\
    \# Transferable utterances  & 6,713 & 7,446 \\
    Avg\# Utterances per dialogue & 10.18 & 9.98 \\
    Avg\# Tokens per utterance  & 17.71 & 21.45 \\
    Std\# Tokens of utterances & 5.31 & 5.29 \\
    \midrule
    Kappa & 0.92  & 0.89 \\
    Agreement  & 93.66 & 91.64 \\
    \bottomrule
    \end{tabular}%
    \caption{Statistics of the datasets we collected}
  \label{tab:statistic}
\end{table}%

A summary of statistics, including Kappa value \cite{snow2008cheap} and inner annotator agreement measured by F-score for both datasets, are shown in Table \ref{tab:statistic}. From the Table \ref{tab:statistic}, we can see that the datasets are highly imbalanced in terms of label distributions and the \textit{normal} utterances occupy the largest proportion. Meanwhile, the length of the utterances content varies greatly. What's more, the distributions of \textit{transferable} utterances' relative locations in the dialogue between the two datasets are significantly different ($p<0.001$ by two-tailed Mann-Whitney U test). From the above observations, it is clear that these two datasets are very challenging for the MHCH task.

\begin{table*}[htbp]
\small
  \centering
  \resizebox{1\textwidth}{!}{
    \begin{tabular}{l|cccccc|cccccc}
    \toprule
    \multicolumn{1}{c|}{\multirow{2}[4]{*}{\textbf{Models}}} & \multicolumn{6}{c|}{\textbf{Clothing}}        & \multicolumn{6}{c}{\textbf{Makeup}} \\
\cmidrule{2-13}          & F1    & Macro F1 & AUC   & GT-I  & GT-II & GT-III & F1    & Macro F1 & AUC   & GT-I  & GT-II & GT-III \\
    \midrule
    \midrule
    HRN   & 57.29 & 73.45 & 86.53 & 62.33 & 71.75 & 76.46 & 58.21 & 74.10  & 87.16 & 62.33 & 72.51 & 78.06 \\
    HAN   & 58.17 & 74.12 & 86.76 & 62.93 & 72.17 & 76.72 & 60.08 & 75.29 & 87.21 & 65.44 & 74.89 & 79.87 \\
    BERT  & 56.03 & 72.93 & 83.78 & 59.27 & 68.07 & 73.10 & 56.98 & 73.25 & 81.65 & 61.45 & 71.01 & 76.48 \\
    CRF-ASN & 57.62 & 73.35 & 84.62 & 61.45 & 72.58 & 77.95 & 56.81 & 73.59 & 83.87 & 63.65 & 74.24 & 79.84 \\
    HBLSTM-CRF & 59.02 & 74.39 & 86.46 & 63.61 & 73.72 & 78.82 & 60.11 & 75.43 & 86.08 & 67.01 & 76.29 & 81.16 \\
    DialogueRNN & 59.04 & 74.34 & \underline{86.94} & 63.10  & 73.75 & 79.02 & 61.33 & 76.07 & 88.45 & 66.34 & 75.98 & 81.15 \\
    CASA  & 59.73 & 74.74 & 86.90  & 64.84 & 74.89 & 79.66 & 60.38 & 75.73 & 87.98 & 67.79 & \underline{76.96} & \underline{81.79} \\
    LSTM + LCA & \underline{61.81} & \underline{76.09} & 85.79 & \underline{66.38} & \underline{76.27} & \underline{81.07} & \underline{62.06} & \underline{76.61} & \underline{88.93} & \underline{67.84} & 76.87 & 81.72 \\
    CESTa & 60.47 & 75.15 & 86.22 & 63.97 & 74.63 & 79.63 & 60.24 & 75.22 & 87.14 & 65.16    & 75.91 & 81.49 \\
    \midrule
    \textbf{DAMI (Our Model)} & \textbf{67.19}$^{**}$ & \textbf{79.44}$^{**}$ & \textbf{91.23}$^{**}$ & \textbf{72.83}$^{**}$ & \textbf{81.27}$^{**}$ & \textbf{85.51}$^{**}$ & \textbf{69.26}$^{**}$ & \textbf{80.90}$^{**}$ & \textbf{92.31}$^{**}$ & \textbf{71.80}$^{**}$ & \textbf{79.60}$^{**}$ & \textbf{84.05}$^{**}$ \\
    \midrule
    \text{ }- Emotion & 66.14 & 78.78 & 90.38 & 67.68 & 77.56 & 82.46 & 65.63 & 78.56 & 90.79 & 67.65 & 77.16 & 82.43 \\
    \text{ }- Matching & 63.69 & 77.67 & 89.56 & 67.92 & 75.90  & 80.35 & 63.76 & 77.37 & 90.18 & 66.16 & 76.24 & 81.80 \\
    \text{ }- Difficulty-Assisted & 65.27 & 78.39 & 90.56 & 68.93 & 76.61 & 80.62 & 64.35 & 78.24 & 90.72 & 68.85 & 76.79 & 81.27 \\
    \text{ }$\hookrightarrow$ + Attention & 66.67 & 79.15 & 91.11 & 69.17 & 76.95 & 81.11 & 67.25 & 79.63 & 91.08 & 69.26 & 78.00    & 82.78 \\
    \bottomrule
    \end{tabular}
    }

    \caption{Performance (\%) on Clothing and Makeup datasets. \underline{\textit{underline}} shows the best performance for baselines. The last four rows show the results of ablation study. ${}^{**}$ indicates statistical significance at $p <$ 0.01 level compared to the best performance of baselines. $\lambda$ of GT-T is 0.}
  \label{tab:compare}%
\end{table*}%

\section{Experiments and Results}
\label{sec:exp}

\subsection{Experimental Settings}
All of the tokenization and part-of-speech tagging are performed by a popular Chinese word segmentation utility called jieba\footnote{https://pypi.org/project/jieba/}. After preprocessing, the datasets are partitioned for training, validation and test with an 80/10/10 split. For all the methods, we apply fine-tuning for the word vectors. The word vectors are initially trained on \textbf{Clothing} and \textbf{Makeup} corpora by using CBOW \cite{mikolov2013efficient}. The dimension of word embedding is set as 200 and the vocabulary size is 19.5K. Other trainable model parameters are initialized by sampling values from the Glorot uniform initializer \cite{glorot2010understanding}. Hyper-parameters of our model and baselines are tuned on the validation set. The sizes of LSTM hidden state $k$, attention units $z$ are all 128, and batch size is set as 128. The dropout \cite{srivastava2014dropout} rate is 0.75 and the number of epochs is set as 30. The $L_2$ regularization weight is $10^{-4}$. Finally, we train the models with an initial learning rate of 0.0075.
All the methods are implemented by Tensorflow\footnote{https://www.tensorflow.org/} and run on a server configured with a Tesla V100 GPU, 8 CPU, and 16G memory.

\subsection{Experimental Results}

We compare our proposed model with the following state-of-the-art dialogue classification models, which mainly come from similar tasks, such as dialogue sentiment classification and dialogue act classification. We briefly describe these baseline models below:
1)
\textbf{HRN} \cite{lin2015hierarchical}: It uses a bidirectional LSTM to encode utterances, which are then fed into a standard LSTM for context representation.
2)
\textbf{HAN} \cite{yang2016hierarchical}: HAN is a hierarchical network, which has two levels of attention mechanisms, applied at the words and utterances.
3)
\textbf{BERT} \cite{kenton2019bert}: It uses a pre-trained BERT model to construct the single utterance representations for classification.
4)
\textbf{CRF-ASN} \cite{chen2018dialogue}: It extends the structured attention network to the linear-chain conditional random field layer, which takes both contextual utterances and corresponding dialogue acts into account.
5)
\textbf{HBLSTM-CRF} \cite{kumar2018dialogue}: It is a hierarchical recurrent neural network using bidirectional LSTM as a base unit and two projection layers to combine utterances and contextual information. 
6)
\textbf{DialogueRNN} \cite{majumder2019dialoguernn}: It is a method based on RNNs that keeps track of the individual party states throughout the conversation and uses the information for emotion classification.
7)
\textbf{CASA} \cite{raheja2019dialogue}: It leverages the effectiveness of a context-aware self-attention mechanism to capture utterance level semantic text representations on prior hierarchical recurrent neural network.
8)
\textbf{LSTM+LCA} \cite{dai2020local}: It is a hierarchical model based on the revised self-attention to capture intra-sentence and inter-sentence information.
9)
\textbf{CESTa} \cite{wang2020contextualized}: It employs LSTM and Transformer to encode context and leverages a CRF layer to learn the emotional consistency in the conversation.

Table \ref{tab:compare} shows the experimental results of the methods on the Clothing and Makeup datasets.
Overall, CESTa, LSTM+LCA, CASA, and DialogueRNN perform better than other baselines, which indicates contextual information is important for the MHCH tasks.
Except for HAN, BERT, and our model, the other models simply use either convolutional or recurrent neural networks to get utterances representations. The experimental results indicate that comprehensive utterances representations have a positive influence on the results (HAN and BERT compare with HRN). 
We will explore how to combine BERT with our proposed components efficiently in the future.
Also, our model utilizes difficulty-assisted encoding to get comprehensive representations. Unlike conventional attention methods, our weights are computed by the combination of word embeddings, term frequencies, position embeddings, and part-of-speech tags to capture the difficulty information associated with utterances. The results show the superiority of our method.

As shown in Table \ref{tab:compare}, our approach achieves the best performance against the strong baselines on the Clothing and Makeup datasets for the MHCH task. We also conduct Wilcoxon signed-rank tests with continuity correction between our method with the baselines (15 runs), and the results show the improvements are significant with ${p < 0.01}$. These experimental results demonstrate the effectiveness of utilizing matching and difficulty information to enhance the hierarchical network capability for the MHCH task. Meanwhile, we also find that, the performance of baseline models will change in different magnitudes with the change of $\lambda$. Taking GT-II on Makeup dataset as an example, LSTM+LCA performs better than CASA when ${\lambda \leq -0.5}$, while it is opposite when ${\lambda \ge -0.25}$. Overall, our models perform reliably better than baseline models\footnote{https://github.com/WeijiaLau/MHCH-DAMI/gtt\_plot.pdf}.

\begin{figure*}[!ht]
    \centering
    \includegraphics[width=17.5cm]{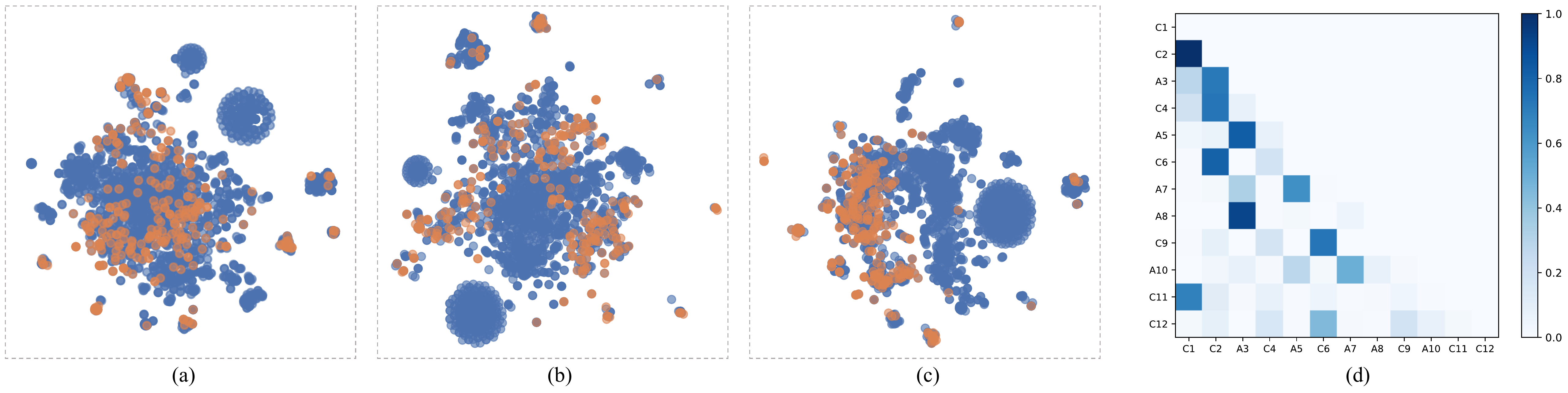}
    \caption{ (a), (b), and (c) are the visualizations of the utterance representation vectors. Specifically, (a) encodes utterances by BiLSTM. (b) encodes utterances by BiLSTM + self-attention. (c) is the full version of Difficulty-Assisted Encoding. \textbf{Orange}/\textbf{Blue} denotes a \textit{transferable}/\textit{normal} utterance; (d) The visualization of matching features $V_M$ for the given example.}
    \label{fig:visual}
\end{figure*}

\subsection{Analysis and Discussion}
\textbf{Ablation study}. To evaluate the contributions of the proposed modules in our model, we implement several model variants for ablation tests respectively. 
\textbf{- Emotion} gets the utterance representation without concatenating the emotional score.
\textbf{- Matching} removes the Matching Inference mechanism. 
\textbf{- Difficulty-Assisted} replaces the Difficult-Assisted Encoding with one layer BiLSTM. 
\textbf{+ Attention} replaces the Difficulty-Assisted Encoding with one layer BiLSTM enhanced by self-attention \cite{yang2016hierarchical}.
The results are recorded in the last four rows of Table \ref{tab:compare}. For the sake of fairness, we keep the other components in the same settings when modifying one module.

We can observe that \textbf{- Emotion} performs well but still worse than DAMI, indicating that concatenating emotional information is helpful but not noticeable compared with other components.
Without \textbf{Matching} or \textbf{Difficulty-Assisted}, the performances drop a lot but are still better than baselines under a few indicators, which proves the effectiveness of these two modules, respectively.

Since our Difficulty-Assisted module is attention-based, we replace it with an attention mechanism. Working with Matching Mechanism, it also outperforms many baselines but not as competent as the full version of DAMI. By removing Matching and Difficulty-Assisted simultaneously, the model degenerates to an HRN-like approach, which performs worst of baselines. This observation indicates that the resonance of these two modules can make essential contributions to the task. 
\begin{figure}[t]
    \centering
    \includegraphics[width=8.4cm]{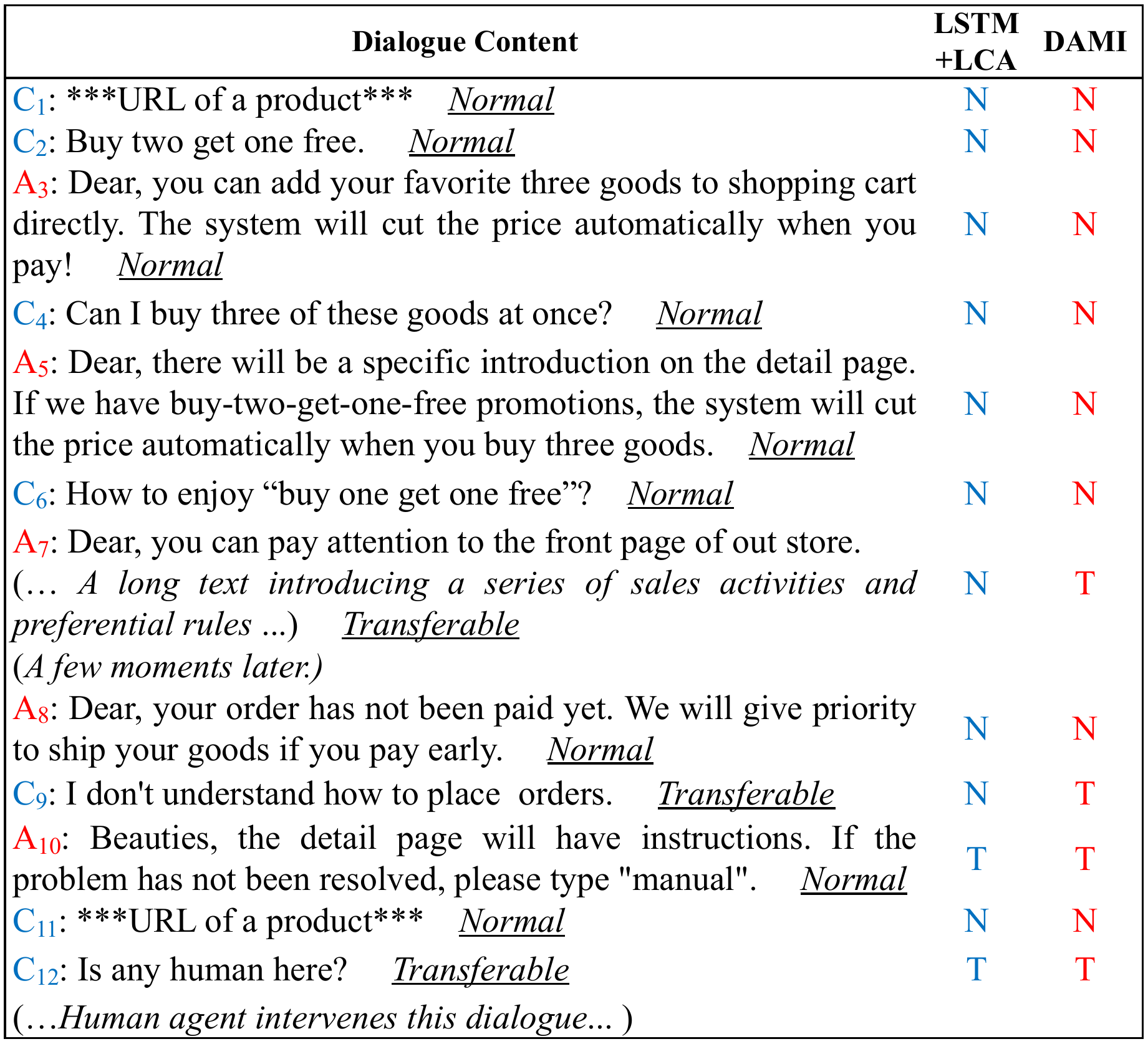}
    \caption{An example from test data. \textbf{C$_i$}/\textbf{A$_i$} denotes Customer/Chatbot utterance, followed by a true \underline{\textit{label}}. The second and third columns are the predictions of LSTM+LCA and DAMI, respectively. N/T denotes \textit{Normal}/\textit{Transferable}.}
    \label{fig:case}
\end{figure}
Figure \ref{fig:visual} (a), (b), and (c) visualize the 3,054 utterance representation vectors $v_t$ from Clothing test dataset by t-SNE \cite{maaten2008visualizing}. Difficulty-Assisted Encoding helps to distinguish the \textit{transferable} utterances from the main clusters of \textit{normal} utterances effectively. Although there are still some overlaps between two types of utterances, they are mainly: (1) Repeated utterances. Due to the incompetence of chatbot, customer or chatbot may repeat semantically similar utterances; (2) Low relevance responses. The chatbot may respond to the customer with a safe but not highly relevant answer, e.g., a universal response. However, those utterances can be correct in perfect matched dialogues. By introducing contextual and matching information, models can distinguish the overlaps.

\textbf{Case Study}. Figure \ref{fig:case} illustrates prediction results with an example which is translated from Chinese text. This case includes three ground-truth \textit{transferable} utterances (A$_7$, C$_9$ and C$_{12}$) and is transferred to the human agent after the 12th round. LSTM+LCA predicts two \textit{transferable} utterances while DAMI predicts four. We combine with the visualization of the matching features in Figure \ref{fig:visual} (d) to explain prediction results.
Considering the context,  A$_{3}$, A$_{5}$, and A$_7$, they are all about placing an order and semantically similar. The terms ``buy one get one free'' in C$_6$ belong to low-frequency terms in corpus, and they affect the chatbot that gives an irrelevant and complicated utterance A$_7$. As a consequence, DAMI correctly predicted A$_7$ as ``\textit{Transferable}''.
In terms of C$_{9}$, along with C$4$ and C$6$, they are all about inquiring activities and placing orders. What's more, the emotion of C$_{9}$ is negative. Our approach captures those matching and emotional features and predicts the utterance as ``\textit{Transferable}'' correctly.
Both LSTM+LCA and DAMI predict A$_{10}$, a universal and indirect response, as ``\textit{Transferable}'' with one round of delay. Finally, the customer complains about the chatbot and asks for a human agent service. Both models recognize this situation and predict correctly.

\section{Conclusion}
\label{sec:conclusion}

In this paper, we introduce the MHCH task to cope with challenging dialogues and enable chatbots' human-algorithm collaboration from a unique perspective. We propose a DAMI network which utilizes a difficulty-assisted encoding module to enhance the representations of utterances. What's more, a matching inference mechanism is introduced for DAMI to capture the contextual matching features. Considering the tolerance property of the MHCH task, a new evaluation metric, namely GT-T, is proposed to assess the model performance. Experimental results on two real-world datasets indicate the efficacy of our model.

In the future, we will investigate personalized information by considering user profiles. Moreover, we will investigate the reverse-handoff task, i.e., handoff from the human agent to machine, when the chatbot has confidence to handle it.

\section{Acknowledgments}
We thank the anonymous reviewers for their helpful comments and suggestions. This work is supported by the National Natural Science Foundation of China (71673211), the National Key R\&D Program of China (2020YFC0832505), and Alibaba Group through Alibaba Research Intern Program and Alibaba Research Fellowship Program.

\bibliography{aaai21}

\end{document}